\pdfoutput=1
\documentclass[10pt,twocolumn,letterpaper]{article}

\usepackage[pagenumbers]{cvpr}

\definecolor{cvprblue}{rgb}{0.21,0.49,0.74}
\usepackage[pagebackref,breaklinks,colorlinks,allcolors=cvprblue]{hyperref}

\title{Out of Sight, Into Memory: Compact Visual Latents Retain Local Evidence in Ultra-High-Resolution Remote Sensing}

\newcommand{\afflogoimg}[1]{\raisebox{-0.36\height}{\includegraphics[height=1.9em]{#1}}}
\newcommand{\afflogo}[1]{%
  \IfFileExists{logos/#1.pdf}{\afflogoimg{logos/#1.pdf}}{%
  \IfFileExists{logos/#1.png}{\afflogoimg{logos/#1.png}}{%
  \raisebox{-0.36\height}{\setlength{\fboxsep}{0pt}\fcolorbox{black!30}{black!6}{\makebox[1.9em][c]{\rule{0pt}{1.9em}\tiny logo}}}}}%
  \hspace{0.3em}}
\newcommand{\affil}[3]{\mbox{\afflogo{#2}$^{#1}$#3}}

\author{%
Meng Tang$^{1}$\qquad
Siyuan Du$^{2}$\qquad
Yuxin Liu$^{3}$\qquad
Shuaishuai Cao$^{4}$\qquad
Di Wang$^{5}$\qquad
Haonan Guo$^{5*}$\\[0.9em]
{\normalsize
\affil{1}{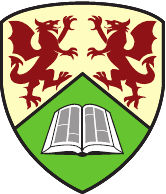}{Aberystwyth University}\qquad
\affil{2}{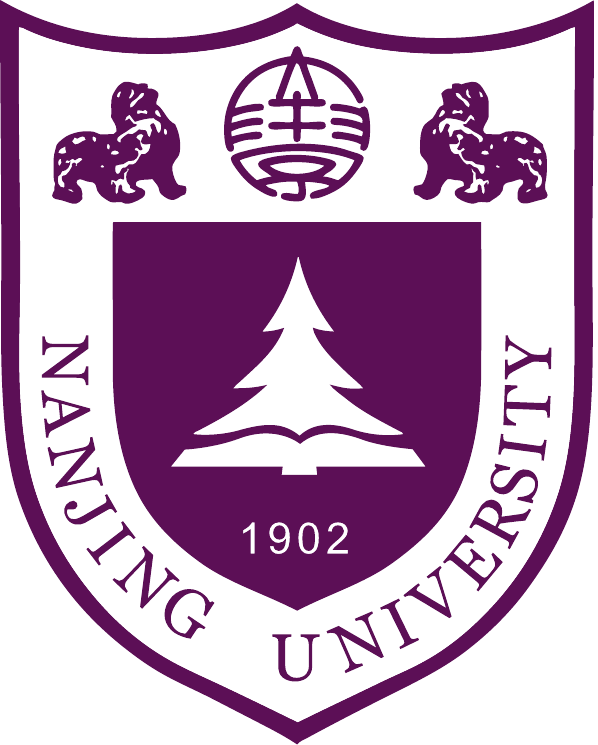}{Nanjing University}\qquad
\affil{3}{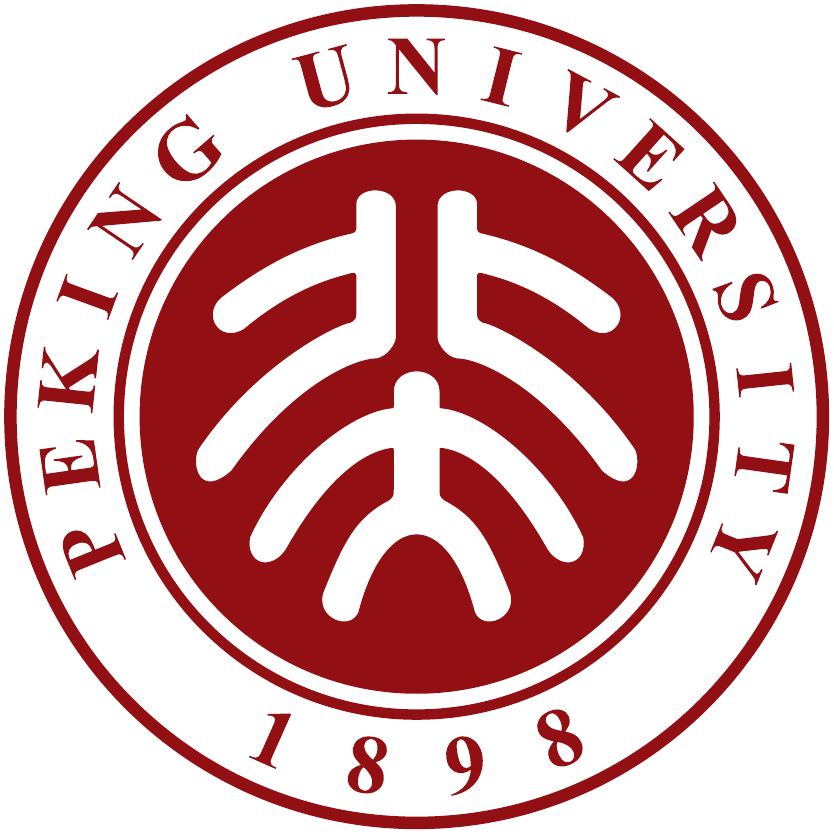}{Peking University}}\\[0.6em]
{\normalsize
\affil{4}{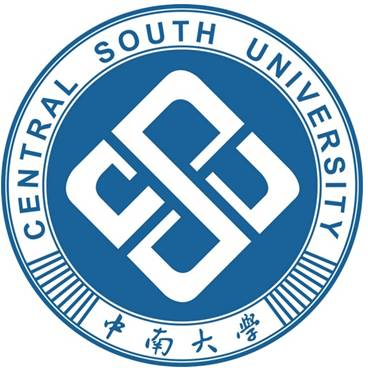}{Central South University}\qquad
\affil{5}{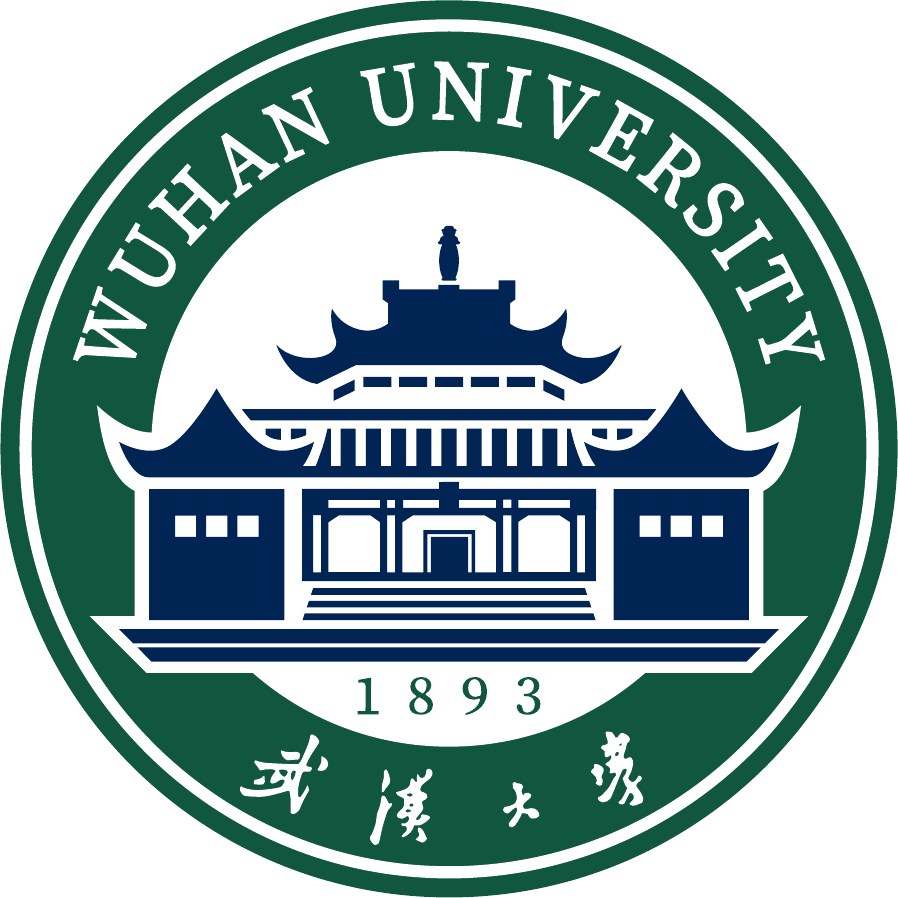}{Wuhan University}}%
}

\hypersetup{
  pdftitle={Out of Sight, Into Memory: Compact Visual Latents Retain Local Evidence in Ultra-High-Resolution Remote Sensing},
  pdfauthor={Meng Tang, Siyuan Du, Yuxin Liu, Shuaishuai Cao, Di Wang, Haonan Guo}
}

\begin{document}
\maketitle
\providecommand{\tbd}[1]{\textcolor{red}{[TBD: #1]}}
\providecommand{\zero}{\textsc{Zero}}
\providecommand{\drop}{\textsc{Drop}}
\providecommand{\swapimg}{\textsc{Swap-Image}}
\providecommand{\swapreg}{\textsc{Swap-Region}}
\providecommand{\finding}[2]{\par\smallskip\noindent\fbox{\parbox{\dimexpr\linewidth-2\fboxsep-2\fboxrule\relax}{\textbf{Finding #1.} #2}}\par\smallskip}

\begin{abstract}
Question answering over ultra-high-resolution (UHR) remote sensing scenes poses a tension for multimodal large language models: global overviews obscure small objects and local attributes, whereas detailed crops consume substantial context. Compressing inspected crops into short sequences of continuous visual latents offers a compact memory once crop tokens are discarded, yet diagnostic studies on natural images find that models often ignore latent content, leaving unclear whether compact latents can retain fine-grained targets lost in downsampled UHR remote sensing scenes. To isolate evidence retention from region selection, we encode high-resolution crops into latent memory tokens for frozen backbones and evaluate both their predictive utility and content dependence. Across two generations of Qwen-VL on LRS-GRO, a 64-token memory achieves 7.6- to 9.9-fold compression of retained crop tokens and improves accuracy by 2.5 to 4.8 percentage points over the downsampled global image alone, recovering most of the high-resolution crop gain with consistent behavior on XLRS-Bench. Substituting memory from another image or another region of the same scene reduces accuracy by 5.8 to 7.4 percentage points and yields lower accuracy than removing memory altogether, demonstrating that predictions depend on the encoded regional evidence. Restoring the source crop within the same model reduces the accuracy loss caused by zeroing or removing memory. These findings establish that compact visual latents retain useful local evidence in UHR remote sensing and identify evidence availability as a key factor governing their contribution.
\end{abstract}

\section{Introduction}
\label{sec:intro}
Ultra-High-Resolution (UHR) Earth observation scenes routinely contain hundreds of millions of pixels, with small objects and fine-grained spatial relationships distributed across vast geographic areas. When multimodal large language models (MLLMs) process a downsampled global overview of such a scene, small targets and local attributes often become indistinguishable. Recent UHR remote sensing methods address this limitation through active perception and selective zooming, inspecting high-resolution crops of task-relevant regions. However, retaining the visual tokens of each inspected crop consumes substantial context budget, creating a bottleneck when reasoning over fine-grained details under limited token capacity \cite{liu2026zoomearth,wang2026geolens}. Compressing an observed crop into a small set of continuous visual latents \cite{li2026latent,wang2026monet,dong2026ilvr,yang2026mirage} and discarding its original tokens offers a natural mechanism for compact visual memory in UHR remote sensing.

Whether MLLMs genuinely read visual evidence stored in such compact latents, however, remains uncertain. Diagnostic studies of latent visual reasoning in general vision domains have shown that replacing latent tokens with uninformative vectors or altering their visual content often has little effect on prediction accuracy \cite{guo2026beyond,li2026imagination,viveiros2026holding}. Such observations indicate that accuracy gains associated with a latent pathway do not necessarily arise from the visual information it carries. At the same time, they point to a key structural distinction between general domain images and UHR remote sensing: evidence availability, defined as whether the task-relevant information carried by a latent can also be recovered from other inputs in the context. When a global image preserves sufficient visual cues, the model retains an alternative route to the answer, consistent with findings that masking input evidence increases reliance on latents \cite{viveiros2026holding}. In UHR remote sensing, downsampling the global scene removes fine-grained local details, leaving compact memory as the sole retained source of zoomed-in visual evidence once crop tokens are discarded.

\begin{figure*}[t]
  \centering
  \makebox[\textwidth][c]{\includegraphics[width=1.05\textwidth]{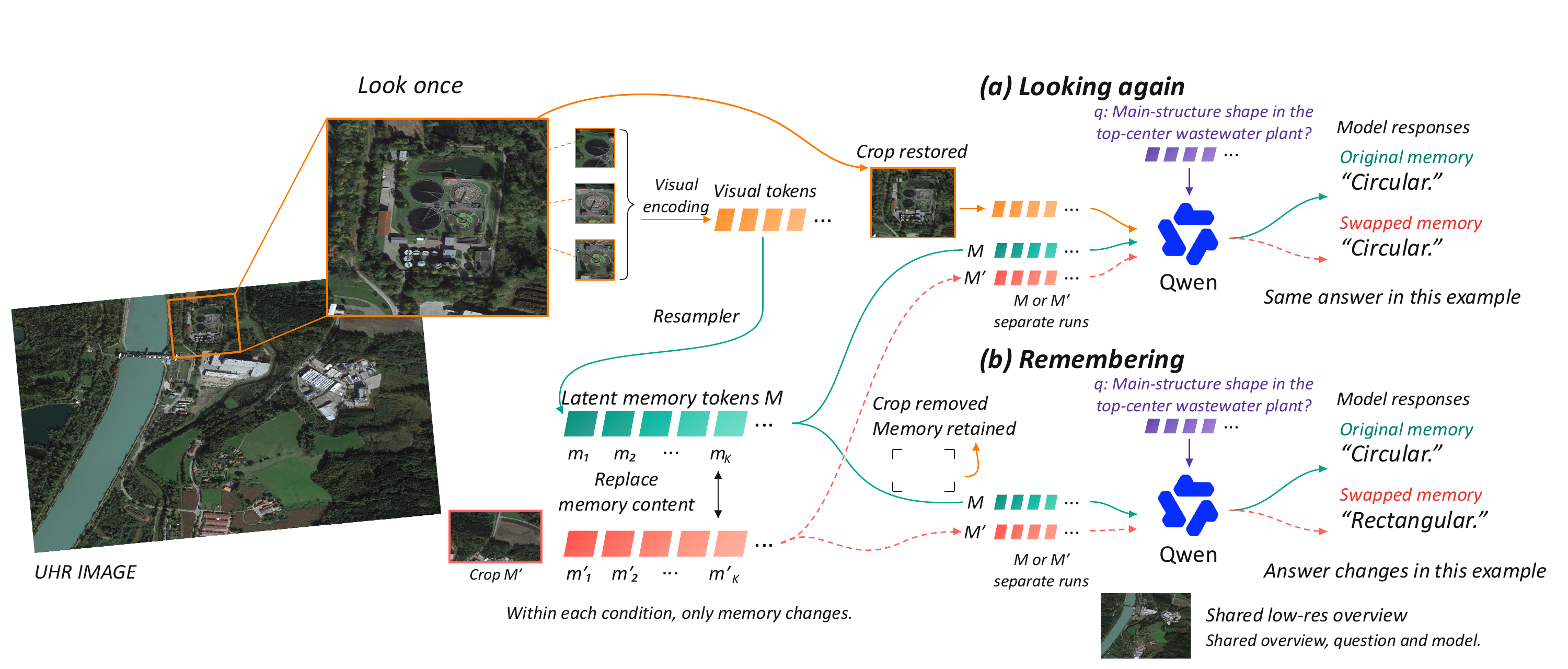}}
  \caption{\textbf{Writing, keeping and probing the memory.}
  A crop of the annotated region is encoded and compressed by a resampler into $K$ memory tokens, after which the crop tokens are discarded.
  In the scarce context the language model sees the question, the thumbnail and the memory, and cannot look again at the crop.
  In the redundant context the crop pixels are added, so the model can look again.
  \zero{}, \drop{}, \swapimg{} and \swapreg{} act on the memory only.}
  \label{fig:method}
\end{figure*}

We investigate whether this natural evidence scarcity enables compact visual latents to serve as an effective, content-dependent memory for UHR question answering. To separate the retention and reading of local evidence from the choice of where to look, we fix crop locations using region annotations (Figure~\ref{fig:method}). Keeping both the vision encoder and the language model frozen, we first calibrate a shared 8-token format prefix on global overviews to absorb answer formatting and label priors across all evaluated conditions, and then train only a lightweight resampler~\cite{jaegle2021perceiver,alayrac2022flamingo} of approximately 10M parameters to map each high-resolution crop into $K$ latent memory tokens. At inference, original crop tokens are removed, and the frozen model answers from the question, the downsampled global image, the shared format prefix, and the $K$ memory tokens.

Across two generations of Qwen-VL \cite{bai2025qwen25vl,bai2025qwen3vl} on the local questions of LRS-GRO \cite{liu2026zoomearth}, compact visual memory delivers substantial context savings alongside consistent accuracy gains. With $K=64$ memory tokens, accuracy improves by 2.5 percentage points on Qwen2.5-VL and 4.8 percentage points on Qwen3-VL relative to using only the downsampled global image. This representation replaces full-resolution crop encodings averaging 634 and 486 tokens, achieving 9.9-fold and 7.6-fold compression of retained crop context while recovering approximately two thirds of the full-crop gain on Qwen2.5-VL and matching full-crop accuracy on Qwen3-VL. It also outperforms matched-budget alternatives on LRS-GRO and transfers directly on Qwen3-VL to XLRS-Bench \cite{wang2025xlrsbench}.

Targeted interventions on the memory further establish that predictions depend on the specific regional evidence it encodes, and that this dependence varies with evidence availability. Holding token count, token positions, and model weights fixed, substituting memory from another image reduces accuracy by 7.4 percentage points on Qwen2.5-VL and 6.2 percentage points on Qwen3-VL, while substituting memory from another annotated region of the same scene reduces accuracy by 6.0 and 5.8 percentage points, respectively. In both models, mismatched memory yields lower accuracy than removing the memory altogether, confirming that the model actively reads the retained local content. Restoring the source crop to the context within the same trained model reduces the accuracy loss caused by zeroing or removing memory by 2.7 to 3.3 percentage points on LRS-GRO, with even larger reductions on XLRS-Bench. Thus, the contribution of the same visual latent shifts according to whether its local evidence is accessible elsewhere in the context.

Our contributions are summarized as follows:
\begin{itemize}
    \item We demonstrate that compact visual latents efficiently retain fine-grained local evidence in UHR remote sensing. Across Qwen2.5-VL and Qwen3-VL on LRS-GRO, a 64-token memory achieves 7.6-fold to 9.9-fold compression of retained crop tokens, improves accuracy by 2.5 to 4.8 percentage points over the global image alone, and recovers most or all of the high-resolution crop gain, with consistent cross-dataset transfer on Qwen3-VL to XLRS-Bench.
    \item We establish causal dependence on encoded regional content through controlled substitutions and matched-budget baselines. Replacing memory with latents from another image or another region within the same UHR scene reduces accuracy by 5.8 to 7.4 percentage points and harms prediction more than removing memory altogether.
    \item We identify evidence availability as a key factor governing visual latent reliance in MLLMs. Restoring the source crop within the same model substantially attenuates the accuracy loss of zeroing or removing memory, explaining why the loss of fine detail in downsampled UHR overviews activates latent memory use.
\end{itemize}
\section{Related Work}
\label{sec:related}

\paragraph{Ultra-High-Resolution Remote Sensing.}
UHR remote sensing requires models to interpret small objects and local attributes distributed across large scenes. Benchmarks such as XLRS-Bench~\cite{wang2025xlrsbench} and LRS-VQA~\cite{luo2025lrsvqa} highlight the difficulty of preserving subtle local details under constrained visual processing budgets. Existing methods improve access to relevant regions through selective processing. Luo et al.~\cite{luo2025lrsvqa} combine a dynamic image pyramid with tile selection guided by text and token pruning, while ZoomEarth~\cite{liu2026zoomearth} and Wang et al.~\cite{wang2026geolens} employ adaptive cropping, zooming, and visual tools to acquire local visual evidence. Whereas these approaches focus on selecting where to inspect a scene, retaining the visual tokens of inspected crops still consumes substantial context. Our work complements region selection by demonstrating that acquired local evidence remains useful when compressed into compact visual latents after the original crop tokens are discarded.

\paragraph{Visual Token Compression and Memory.}
Visual token compression produces compact representations of visual inputs to reduce computational and context costs. Perceiver~\cite{jaegle2021perceiver} aggregates dense inputs into a latent array, while Flamingo~\cite{alayrac2022flamingo} uses a Perceiver Resampler to produce a fixed number of visual tokens. Token Merging~\cite{bolya2023tome} combines similar tokens in vision transformers, and FastV~\cite{chen2024fastv} prunes redundant visual tokens within multimodal language models. Related work on visual memory, including MA-LMM~\cite{he2024malmm} and MovieChat~\cite{song2024moviechat}, retains information from earlier frames to support the understanding of long videos. In our setting, a resampler with learned queries encodes crops at high resolution into compact latent memories. Beyond evaluating accuracy under reduced token budgets, we examine whether model predictions genuinely depend on the regional visual information preserved in these tokens.

\paragraph{Latent Visual Reasoning and Diagnostics.}
Latent visual reasoning introduces continuous tokens as intermediate visual representations within multimodal language models~\cite{li2026latent,dong2026ilvr,yang2026mirage}. Methods such as Monet~\cite{wang2026monet} train models to generate these representations through visual supervision and distillation to support reasoning in latent space. Alongside advances in latent construction, diagnostic studies have examined whether model predictions depend on the visual content of these representations. Li et al.~\cite{li2026imagination} identify weak connections between input images, latent representations, and final predictions in evaluated models. Guo et al.~\cite{guo2026beyond} distinguish the content of latent slots from boundary markers and formatting, showing that substantial performance gains can persist without the slot content. Viveiros et al.~\cite{viveiros2026holding} connect latent bypass to visual evidence already accessible in the input and demonstrate increased reliance when relevant image regions are masked during training. CVRR~\cite{park2026cvrr} makes the recurrent latent state the required visual pathway for prediction and shows that restoring an alternative visual route partially compensates for corrupted latent content. These findings indicate that the contribution of visual latents depends on both their encoded content and the surrounding visual context. Extending this perspective to UHR remote sensing, we show that compact latent memories preserve local details obscured in the downsampled global overview and that model predictions rely on this regional content. Furthermore, restoring the source crop within the same model reduces reliance on memory, supporting evidence availability as an explanation for the contribution of visual latents.

\section{Visual Memory Testbed and Protocol}
\label{sec:setting}
\label{sec:method}

To evaluate whether compact visual latents preserve useful local evidence in UHR remote sensing, we construct a controlled question answering testbed. This section details the memory construction pipeline, the two context conditions and causal interventions used to probe content reliance, and the evaluation protocol.

\subsection{Compact Crop Memory Construction}
\label{sec:writing}

Given a question $q$ about a UHR scene, the model receives a downsampled global image resized to a maximum side length of 1,024 pixels. Although this overview preserves global scene layout, small targets and local attributes often become indistinguishable at this scale. To supply fine local detail while separating memory retention from region localization, we extract a crop from the annotated region associated with each question. Following the LRS-GRO implementation~\cite{liu2026zoomearth}, we enlarge the annotated bounding box by 50\%, enforce a minimum side length of 512 pixels at native resolution, and resize the crop so that its longer side does not exceed 1,024 pixels.

The frozen vision encoder maps the crop to a sequence of visual features, which are compressed into a memory $m$ of $K$ continuous tokens in the embedding space of the language model (Figure~\ref{fig:method}). Our resampler~\cite{jaegle2021perceiver,alayrac2022flamingo} consists of 2 layers with internal dimension 512, 8 attention heads, feedforward width 2,048, and 64 learned queries; for any budget $K \in \{4, 8, 16, 32, 64\}$ ($K=64$ by default), it selects the first $K$ queries. Each layer applies cross attention from the queries to the crop features followed by a feedforward network, projecting back to the hidden dimension of the language model (3,584 on Qwen2.5-VL, totaling 10.0M trainable parameters). Its output projection is initialized to zero and added as a residual to spatially pooled crop features, so the module reduces exactly to the Average Pooling baseline before training.

Throughout training, both the vision encoder and the language model remain strictly frozen without LoRA~\cite{hu2022lora} or backbone tuning. Using 2,117 local training questions from 228 scenes and 223 validation questions from 54 scenes in LRS-GRO, we run two stages with AdamW, batch size 1 with 8 gradient accumulation steps, gradient clipping at 1.0, and a constant learning rate without warmup. To prevent trained memory from winning merely by adapting output formatting or answer priors on a frozen backbone, Stage 1 initializes an 8-token format prefix (28.7K parameters on Qwen2.5-VL and 32.8K on Qwen3-VL) from the first 8 token embeddings of ``Answer with a short lowercase phrase and no punctuation.'' and trains it for 3 epochs at a learning rate of $3\times 10^{-4}$ with zero weight decay using answer cross entropy on global overviews and questions alone, selecting the checkpoint with lowest validation cross entropy. This 8-token format prefix is then frozen and inserted after the visual or memory tokens and before the question in every evaluated condition and intervention. Stage 2 updates only the resampler (10.0M parameters on Qwen2.5-VL and 10.5M on Qwen3-VL) for 4 epochs at a learning rate of $1\times 10^{-4}$ with weight decay 0.01, minimizing answer cross entropy combined with a counterfactual margin loss and logit distillation over the top 64 tokens from uncompressed crop inputs.

\subsection{Diagnostic Contexts and Interventions}
\label{sec:availability_def}
\label{sec:scarcity}
\label{sec:hypotheses}

Let $c$ denote the answering context accompanying the memory $m$. We distinguish overall predictive utility from causal reliance on memory content. Predictive utility is measured by comparing accuracy against a baseline that receives only the question and the downsampled global overview. Reliance is measured within the same trained model by intervening on $m$ while holding $c$ and all model weights fixed. For an intervention that replaces $m$ with $\tilde{m}$, we define paired reliance in percentage points on identical questions as:
\begin{equation}
\mathrm{Rel} = \mathrm{Acc}_{c,\,m} - \mathrm{Acc}_{c,\,\tilde{m}} .
\label{eq:rel}
\end{equation}
A positive value of $\mathrm{Rel}$ indicates that corrupting or removing the memory reduces prediction accuracy.

Evidence availability refers to whether the local information carried by $m$ can also be recovered from the surrounding context $c$. We evaluate how evidence availability shapes memory reliance by comparing two answering contexts at inference time (Figure~\ref{fig:method}):
\begin{itemize}
    \item \textbf{Scarce context (Remembering).} Original crop tokens are discarded, so $c$ contains only the question and the downsampled global overview. Because the model cannot inspect the crop pixels again, $m$ serves as the sole retained carrier of zoomed local evidence.
    \item \textbf{Redundant context (Looking again).} Original crop tokens from which $m$ was written are restored to $c$ alongside the same memory $m$. Because the zoomed evidence is simultaneously accessible as pixels, comparing reliance across the two contexts while keeping model weights, questions, and memories fixed isolates the effect of evidence availability within a single model, providing an inference counterpart to the masked training analysis of Viveiros~\etal~\cite{viveiros2026holding}.
\end{itemize}

We apply four targeted interventions to $m$. \zero{} replaces every memory token with a zero vector, and \drop{} removes the memory tokens from the input sequence; together they test whether the retained pathway contributes necessary information. \swapimg{} replaces $m$ with memory encoded from a different scene, whereas \swapreg{} replaces $m$ with memory encoded from another annotated region within the same scene. Because \swapimg{} and \swapreg{} preserve the exact number, sequence positions, and feature distribution of the memory tokens as well as the trained pathway that consumes them, they isolate dependence on regional visual content from potential artifacts caused by sequence truncation.

\subsection{Benchmarks, Controls, and Scoring Protocol}
\label{sec:protocol}

We conduct our primary evaluation on LRS-GRO~\cite{liu2026zoomearth}, covering 7,393 local test questions across 839 UHR scenes using Qwen2.5-VL-7B-Instruct~\cite{bai2025qwen25vl} and Qwen3-VL-8B-Instruct~\cite{bai2025qwen3vl}, which tokenize visual inputs at 28 and 32 pixels per token, respectively. To test generalization across datasets, we evaluate the Qwen3-VL model trained on LRS-GRO directly without further training on 1,657 questions with reference boxes from 856 scenes in XLRS-Bench~\cite{wang2025xlrsbench}, denoted XLRS-Bench-lite, applying the identical crop extraction rule to each reference box.

All evaluated conditions share the frozen backbone, question, downsampled global overview, and frozen 8-token Stage 1 format prefix. \textbf{Thumbnail}, \textbf{High-Res Crop}, \textbf{Low-Res Pixels}, \textbf{Average Pooling}, and \textbf{Text Notes} introduce zero additional trainable parameters beyond this shared prefix: Thumbnail provides no crop, High-Res Crop supplies uncompressed crop tokens, and the three compact baselines at budget $K$ compress the crop via image downsampling, spatial feature pooling (the exact zero-parameter initialization of \textbf{Memory}), or truncated descriptions generated by the frozen backbone without the prefix (at most $K$ tokens). \textbf{Soft Prompt} places $K$ static learned vectors independent of the crop at the memory token positions using answer cross entropy; because it has far fewer parameters than the resampler, it tests adding $K$ learned continuous slots without crop input, whereas full resampler parameter capacity is controlled in Section~\ref{sec:dependence} via \zero{} and \swapimg{} using the identical Stage 2 checkpoint. Across tables, reported token counts measure only the regional evidence payload per question, excluding the global overview, the shared 8-token format prefix, and region label text.

All runs use greedy decoding under a fixed generation configuration and share a deterministic scoring protocol. When a question specifies candidate answers, we score the likelihood assigned to each candidate rather than parsing generated text; otherwise, we apply official benchmark matching rules. Likelihood scoring prevents output formatting failures from confounding memory interventions: in a pilot on XLRS-Bench using generative extraction, \zero{} lowered accuracy by 19.8 percentage points largely through invalid output formats. All comparisons are paired on identical test questions, which remain strictly excluded from model development and checkpoint selection. We report 95\% confidence intervals from a cluster bootstrap over scenes to account for correlation among questions sharing the same scene, considering an intervention effect statistically supported when the lower bound of its 95\% confidence interval exceeds zero.
\section{Empirical Findings}
\label{sec:results}

We organize our empirical analysis around three progressive questions: whether compact visual memory preserves fine local details under tight token budgets, whether predictions genuinely depend on the regional visual content stored in memory, and why UHR remote sensing activates latent reading where general vision settings often exhibit bypass. Unless stated otherwise, results use the default budget of $K=64$ tokens.

\subsection{Compact Memory Preserves Local UHR Evidence}
\label{sec:utility}

\begin{table*}[t]
\centering
\small
\setlength{\tabcolsep}{7pt}
\renewcommand{\arraystretch}{1.12}
\begin{tabular*}{\textwidth}{@{\extracolsep{\fill}}lrrrrrr@{}}
\toprule
& \multicolumn{4}{c}{LRS-GRO}
& \multicolumn{2}{c}{XLRS-Bench-lite} \\
\cmidrule(lr){2-5}\cmidrule(l){6-7}
& \multicolumn{2}{c}{Qwen2.5-VL}
& \multicolumn{2}{c}{Qwen3-VL}
& \multicolumn{2}{c}{Qwen3-VL} \\
\cmidrule(lr){2-3}\cmidrule(lr){4-5}\cmidrule(l){6-7}
Condition & Tokens & Acc. & Tokens & Acc. & Tokens & Acc. \\
\midrule
Thumbnail       & 0   & 54.6 & 0    & 57.2 & 0     & 36.3 \\
Soft Prompt     & 64  & 53.8 & 64   & 57.7 & 64    & 37.5 \\
Low-Res Pixels  & 64  & 55.8 & 64   & 59.8 & 64    & \textbf{52.9} \\
Text Notes      & 47  & 54.9 & 46.5 & 58.0 & 49.4  & 39.4 \\
Average Pooling & 64  & 54.9 & 64   & 58.7 & 64    & 43.6 \\
\textbf{Memory} & 64  & \textbf{57.1} & 64 & \textbf{62.0} & 64 & 48.5 \\
\midrule
High-Res Crop   & 634 & 58.4 & 486  & 61.6 & 321.8 & 53.7 \\
\bottomrule
\end{tabular*}
\caption{\textbf{Predictive utility and retained context cost.}
Accuracy (\%) on LRS-GRO (7,393 questions from 839 scenes) and XLRS-Bench-lite (1,657 questions from 856 scenes).
All conditions include the question, the downsampled global overview, and the shared frozen 8-token Stage 1 format prefix.
Compact representations use a budget of $K=64$ tokens, whereas High-Res Crop retains the uncompressed crop tokens as an upper reference.
Tokens counts only the regional evidence payload per question beyond this shared context (excluding global overview tokens, the 8-token format prefix, and region label text).
Bold values mark the highest accuracy within the budget of $K=64$ tokens.}
\label{tab:memory_utility}
\end{table*}

\begin{figure}[t]
  \centering
  \includegraphics[width=\columnwidth]{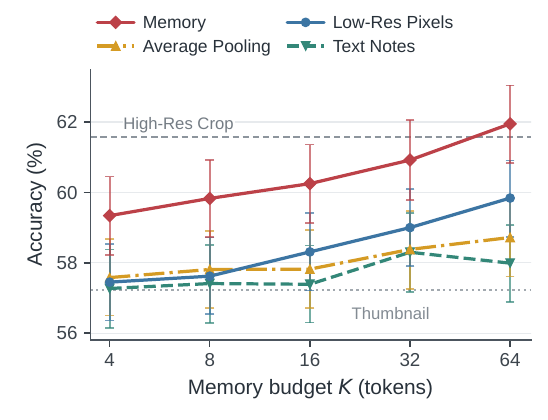}
  \caption{\textbf{Accuracy across token budgets on LRS-GRO.}
  Accuracy of Memory, Low-Res Pixels, Text Notes, and Average Pooling on Qwen3-VL in the scarce context for $K \in \{4,8,16,32,64\}$.
  Horizontal reference lines indicate Thumbnail and High-Res Crop.
  Error bars show 95\% confidence intervals from a cluster bootstrap over scenes.}
  \label{fig:budget}
\end{figure}

Table~\ref{tab:memory_utility} evaluates predictive utility and retained context cost when original crop tokens are discarded. On LRS-GRO, adding $K=64$ latent memory tokens raises accuracy over Thumbnail from 54.6\% to 57.1\% (+2.5 percentage points) on Qwen2.5-VL and from 57.2\% to 62.0\% (+4.8 percentage points) on Qwen3-VL. These 64 tokens replace full crop representations averaging 634 tokens on Qwen2.5-VL and 486 tokens on Qwen3-VL, compressing retained crop context by factors of 9.9 and 7.6, respectively. Despite this compression, Memory recovers roughly two thirds of the full crop improvement on Qwen2.5-VL (57.1\% vs.\ 58.4\%) and matches High-Res Crop on Qwen3-VL (62.0\% vs.\ 61.6\%, a paired difference of $0.4\;[-0.4,\,1.2]$ points). When transferred directly to XLRS-Bench-lite without additional training, Memory on Qwen3-VL improves accuracy over Thumbnail by 12.2 percentage points (from 36.3\% to 48.5\%), preserving approximately 70\% of the High-Res Crop gain while compressing an average of 321.8 crop tokens by a factor of 5.0.

Comparisons against controls at matched budgets confirm that these improvements cannot be explained by adding static learned token slots without crop input. Soft Prompt yields 53.8\% on Qwen2.5-VL and 57.7\% on Qwen3-VL on LRS-GRO, and 37.5\% on XLRS-Bench-lite, remaining near Thumbnail across all settings (while Section~\ref{sec:dependence} rules out the 10.0M to 10.5M resampler parameters by evaluating the identical checkpoint under \zero{} and \swapimg{}).

\paragraph{Finding 1.}
\textit{Compact visual latents retain local UHR evidence at a fraction of the original crop token cost, recovering most or all of the accuracy gain of uncompressed crops and outperforming alternative representations across budgets on the training domain.}

\subsection{Predictions Depend on Regional Visual Content}
\label{sec:dependence}

\begin{table*}[t]
\centering
\small
\setlength{\tabcolsep}{3.5pt}
\renewcommand{\arraystretch}{1.12}
\begin{tabular*}{\textwidth}{@{\extracolsep{\fill}}lrc r c r c@{}}
\toprule
& \multicolumn{4}{c}{LRS-GRO}
& \multicolumn{2}{c}{XLRS-Bench-lite} \\
\cmidrule(lr){2-5}\cmidrule(l){6-7}
& \multicolumn{2}{c}{Qwen2.5-VL}
& \multicolumn{2}{c}{Qwen3-VL}
& \multicolumn{2}{c}{Qwen3-VL} \\
\cmidrule(lr){2-3}\cmidrule(lr){4-5}\cmidrule(l){6-7}
Condition & Acc. & Rel & Acc. & Rel & Acc. & Rel \\
\midrule
Memory
& 57.1 & 0.0
& 62.0 & 0.0
& 48.5 & 0.0 \\
\textsc{Zero}
& 52.6 & $4.5\;[3.6,\,5.3]$
& 56.7 & $5.2\;[4.4,\,6.0]$
& 36.7 & $11.8\;[9.2,\,14.3]$ \\
\textsc{Drop}
& 53.6 & $3.5\;[2.6,\,4.4]$
& 57.3 & $4.6\;[3.9,\,5.4]$
& 36.6 & $11.8\;[9.3,\,14.4]$ \\
\textsc{Swap-Image}
& 49.7 & $\mathbf{7.4}\;[6.5,\,8.3]$
& 55.8 & $\mathbf{6.2}\;[5.4,\,7.0]$
& 32.3 & $\mathbf{16.1}\;[13.5,\,18.7]$ \\
\addlinespace[2pt]
\textsc{Swap-Region}$^\dagger$
& 50.7 & $\mathbf{6.0}\;[4.9,\,7.1]$
& 56.1 & $\mathbf{5.8}\;[4.9,\,6.7]$
& 32.0 & $\mathbf{16.3}\;[13.5,\,19.3]$ \\
\bottomrule
\end{tabular*}
\caption{\textbf{Dependence on memory content.}
Accuracy (\%) and paired reliance $\mathrm{Rel}$ (percentage points) at $K=64$ in the scarce context.
$\mathrm{Rel}$ equals the accuracy of the matched Memory reference minus the intervention accuracy, with 95\% confidence intervals from a cluster bootstrap over scenes in brackets.
Results on the complete evaluation set use the samples in Table~\ref{tab:memory_utility}.
$^\dagger$\textsc{Swap-Region} uses a matched Memory reference on the subset of scenes containing multiple annotated regions: 5,295 questions from 514 scenes on LRS-GRO and 1,212 questions from 446 scenes on XLRS-Bench-lite, where reference Memory accuracy on the XLRS-Bench-lite subset is 48.3\%.
Bold $\mathrm{Rel}$ values highlight interventions that substitute memory content.
Paired differences are computed prior to rounding.}
\label{tab:memory_reliance}
\end{table*}

To verify that accuracy gains arise from reading visual information stored in memory rather than from structural formatting cues, Table~\ref{tab:memory_reliance} measures paired reliance under targeted interventions in the scarce context. Eliminating memory information consistently degrades performance: on LRS-GRO, \zero{} lowers accuracy by $4.5\;[3.6,\,5.3]$ percentage points on Qwen2.5-VL and $5.2\;[4.4,\,6.0]$ points on Qwen3-VL, while \drop{} reduces accuracy by $3.5\;[2.6,\,4.4]$ and $4.6\;[3.9,\,5.4]$ points, respectively. On XLRS-Bench-lite, both \zero{} and \drop{} reduce Qwen3-VL accuracy by 11.8 percentage points, returning performance to the level of Thumbnail alone (36.6\% to 36.7\% vs.\ 36.3\%).

Content substitutions provide a stricter test by preserving token count, sequence positions, and latent distributions within the identical trained model while altering only the encoded region. Replacing memory with latents from another scene (\swapimg{}) reduces accuracy by $7.4\;[6.5,\,8.3]$ points on Qwen2.5-VL and $6.2\;[5.4,\,7.0]$ points on Qwen3-VL on LRS-GRO, and by $16.1\;[13.5,\,18.7]$ points on XLRS-Bench-lite. Even when global scene context is shared, replacing memory with latents from another annotated region of the same UHR image (\swapreg{}) lowers accuracy by $6.0\;[4.9,\,7.1]$ and $5.8\;[4.9,\,6.7]$ points on LRS-GRO, and by $16.3\;[13.5,\,19.3]$ points on XLRS-Bench-lite. Across every model and dataset, both \swapimg{} and \swapreg{} produce lower accuracy than removing memory altogether (\drop{}), pushing predictions below the Thumbnail baseline. Incorrect regional latents actively mislead the model, demonstrating that predictions rely on the specific local visual evidence encoded in memory.

\paragraph{Finding 2.}
\textit{Model predictions causally depend on the regional visual content of latent memory: substituting latents from another image or another region within the same UHR scene degrades accuracy even more than removing memory altogether.}

\subsection{Evidence Scarcity Activates Latent Reading}
\label{sec:availability}

\begin{table*}[t]
\centering
\resizebox{\textwidth}{!}{%
\begin{tabular}{lll}
\toprule
Intervention & General domain & Ours (LRS-GRO)\\
\midrule
Content of another image & Change between $-4$ and $+2$ points across six model versions on V$^*$ and BLINK~\cite{viveiros2026holding} & $-7.4$ and $-6.2$ points \\
Uninformative content & Accuracy changes by at most 2 points under an intervention with uninformative tokens for each model and dataset~\cite{viveiros2026holding} & $-4.5$ and $-5.2$ points under \zero{} \\
Content of the same image & Answer follows the donor in 5.3\% to 12.7\% of cases, below the 25\% chance level~\cite{guo2026beyond} & $-6.0$ and $-5.8$ points \\
\bottomrule
\end{tabular}%
}
\caption{\textbf{Contextualizing our findings alongside prior latent diagnostics.}
Results for the general domain are taken from the cited studies.
Our values report accuracy changes in percentage points relative to Memory on LRS-GRO for Qwen2.5-VL and Qwen3-VL, respectively.
Because models, benchmarks, and latent sources differ, only the direction and qualitative magnitude of the effect are compared.}
\label{tab:contrast}
\end{table*}

\begin{table*}[t]
\centering
\small
\setlength{\tabcolsep}{5pt}
\renewcommand{\arraystretch}{1.12}
\begin{tabular*}{\textwidth}{@{\extracolsep{\fill}}lllccc@{}}
\toprule
Dataset & Model & Intervention
& Scarce & Redundant & Reduction \\
\midrule
LRS-GRO & Qwen2.5-VL & \textsc{Zero}
& $4.5\;[3.6,\,5.3]$
& $1.2\;[0.5,\,1.9]$
& $\mathbf{3.2}\;[2.4,\,4.1]$ \\
& & \textsc{Drop}
& $3.5\;[2.6,\,4.4]$
& $0.2\;[-0.5,\,0.9]$
& $\mathbf{3.3}\;[2.4,\,4.2]$ \\
& & \textsc{Swap-Region}$^\dagger$
& $6.0\;[4.9,\,7.1]$
& $3.7\;[2.8,\,4.7]$
& $2.3\;[1.4,\,3.3]$ \\
\addlinespace[3pt]
& Qwen3-VL & \textsc{Zero}
& $5.2\;[4.4,\,6.0]$
& $2.5\;[1.8,\,3.1]$
& $\mathbf{2.7}\;[2.0,\,3.5]$ \\
& & \textsc{Drop}
& $4.6\;[3.9,\,5.4]$
& $1.7\;[1.1,\,2.3]$
& $\mathbf{2.9}\;[2.1,\,3.8]$ \\
& & \textsc{Swap-Region}$^\dagger$
& $5.8\;[4.9,\,6.7]$
& $2.1\;[1.4,\,2.8]$
& $3.7\;[2.8,\,4.6]$ \\
\midrule
XLRS-Bench-lite & Qwen3-VL & \textsc{Zero}
& $11.8\;[9.2,\,14.3]$
& $\mathbf{-3.1}\;[-5.4,\,-0.9]$
& $\mathbf{14.9}\;[12.2,\,17.6]$ \\
& & \textsc{Drop}
& $11.8\;[9.3,\,14.4]$
& $\mathbf{-7.2}\;[-9.3,\,-5.1]$
& $\mathbf{19.0}\;[16.3,\,21.7]$ \\
& & \textsc{Swap-Region}$^\dagger$
& $16.3\;[13.5,\,19.3]$
& $16.1\;[13.3,\,18.9]$
& $0.2\;[-2.1,\,2.7]$ \\
\bottomrule
\end{tabular*}
\caption{\textbf{Memory reliance decreases when source evidence is restored.}
Paired reliance $\mathrm{Rel}$ in percentage points at $K=64$.
The scarce context contains the question, global overview, and memory; the redundant context additionally restores the source crop.
Reduction equals $\mathrm{Rel}_{\mathrm{scarce}} - \mathrm{Rel}_{\mathrm{redundant}}$, computed from unrounded paired results, with 95\% confidence intervals from a cluster bootstrap over scenes in brackets.
Negative $\mathrm{Rel}$ indicates that the intervention increases accuracy.
Bold values highlight reductions under \textsc{Zero} and \textsc{Drop} as well as negative reliance in the redundant context.
$^\dagger$\textsc{Swap-Region} uses the subsets specified in Table~\ref{tab:memory_reliance} with matched questions across both contexts.}
\label{tab:evidence_availability}
\end{table*}

The strong content sensitivity observed above contrasts sharply with diagnostic findings in general vision domains. As summarized in Table~\ref{tab:contrast}, substituting latents from another image or injecting uninformative tokens on standard visual benchmarks typically alters accuracy by at most two to four percentage points~\cite{viveiros2026holding}, and donor latents from the same image rarely steer model predictions~\cite{guo2026beyond}. In our scarce UHR setting, analogous interventions consistently induce drops of 4.5 to 7.4 percentage points on LRS-GRO and up to 16.3 points on XLRS-Bench-lite. Unlike natural images where global views often retain sufficient cues to answer local queries, downsampling a UHR Earth observation scene removes small targets and subtle attributes, leaving compact memory as the only carrier of local visual evidence.

To test whether evidence availability directly governs this behavior within the same model, we restore the source crop tokens alongside the memory at inference time (the redundant context) while keeping all model weights, questions, global overviews, and memories unchanged (Table~\ref{tab:evidence_availability}). On LRS-GRO, restoring the crop reduces accuracy loss under \zero{} by $3.2\;[2.4,\,4.1]$ percentage points on Qwen2.5-VL and $2.7\;[2.0,\,3.5]$ points on Qwen3-VL, and reduces loss under \drop{} by $3.3\;[2.4,\,4.2]$ and $2.9\;[2.1,\,3.8]$ points, respectively. On XLRS-Bench-lite, restoring the source crop on Qwen3-VL reduces reliance under \zero{} and \drop{} by $14.9\;[12.2,\,17.6]$ and $19.0\;[16.3,\,21.7]$ percentage points; in this redundant setting, removing the transferred memory (\drop{}) raises accuracy by 7.2 points. Providing direct pixel access to the local region thus supplies an alternative route to the visual evidence and substantially lowers the cost of zeroing or removing memory.

At the same time, access to source pixels does not make the model impervious to conflicting memory content. In the redundant context on LRS-GRO, \swapreg{} still lowers accuracy by $3.7\;[2.8,\,4.7]$ points on Qwen2.5-VL and $2.1\;[1.4,\,2.8]$ points on Qwen3-VL, while on XLRS-Bench-lite reliance under \swapreg{} remains virtually unchanged at $16.1\;[13.3,\,18.9]$ points (a reduction of $0.2\;[-2.1,\,2.7]$ points). Restoring source pixels therefore diminishes the necessity of memory when it is absent, while predictions remain sensitive to mismatched regional content.

\paragraph{Finding 3.}
\textit{Evidence availability directly modulates reliance on visual latents within a single model: restoring high-resolution crop pixels substantially reduces the cost of zeroing or removing memory, explaining why natural detail loss in downsampled UHR overviews activates latent reading.}
\section{Discussion}
\label{sec:discussion}

\paragraph{Visual Memory and Evidence Reconciliation.}
Our experiments establish that compact visual latents written from inspected crops serve as genuine local memory in UHR remote sensing. After original crop tokens are removed, retaining memory improves accuracy, and substituting its regional content degrades predictions across both benchmarks. Pairing predictive utility with causal content sensitivity addresses a central concern raised by Guo et al.~\cite{guo2026beyond}, who show that accuracy gains from latent tokens need not imply that their contents store visual information. Complementing diagnostics of latents generated as intermediate reasoning states, our results confirm a concrete memory function when latents encode observed local crops obscured in the global overview. Restoring the source crop further links this function to surrounding visual context, extending the account of Viveiros et al.~\cite{viveiros2026holding} to retained crop representations while uncovering an operational asymmetry: access to source pixels reduces the penalty of zeroing or dropping memory, yet does not protect predictions against mismatched regional latents (\swapreg{}), particularly on XLRS-Bench-lite (Table~\ref{tab:evidence_availability}). For sequential remote sensing agents, deciding when to retain past crop memories and verifying their consistency with newly inspected pixels therefore represent distinct design requirements.

\paragraph{Representation Choices Across Domains and Tasks.}
Comparisons across datasets and question types clarify where learned query resampling is most effective. On XLRS-Bench-lite, 64 memory tokens on Qwen3-VL preserve roughly 70\% of the accuracy gain provided by High-Res Crop over Thumbnail at fivefold token compression (Table~\ref{tab:memory_utility}) while exhibiting strong content dependence (Table~\ref{tab:memory_reliance}). Under the same budget of 64 tokens, however, Low-Res Pixels achieves higher accuracy without relying on resampling queries learned on LRS-GRO, indicating that the relative advantage of learned compression over spatial downsampling varies under domain shift. Task granularity introduces a related distinction on LRS-GRO: although Qwen3-VL with Memory matches High-Res Crop in overall accuracy, its accuracy on object counting questions drops to 32.1\% compared with 41.0\% for High-Res Crop. Learned query resampling effectively condenses regional semantics and attributes into a small token budget, whereas tasks requiring discrete instance enumeration or direct transfer across imagery distributions benefit from preserving explicit spatial structure.

\paragraph{Integration with Active Perception and Scope.}
Fixing crop locations through region annotations or reference boxes separates the retention and reading of local evidence from localization errors. Combining compact latent memory with learned region selection~\cite{liu2026zoomearth,wang2026geolens} to support reasoning across multiple inspected regions is a natural next step for UHR Earth observation systems. Evaluating architectures beyond two generations of the Qwen-VL family will also clarify how broadly these behaviors hold across multimodal backbones. Finally, our compression ratios measure the visual context retained within the language model; because the vision encoder still processes each crop prior to resampling, benchmarking overall inference latency remains an important direction for deployment.
\section{Conclusion}
\label{sec:conclusion}

Reasoning over UHR remote sensing imagery under constrained token budgets requires preserving fine local details that disappear in downsampled global overviews without retaining hundreds of tokens per inspected crop. Across two generations of Qwen-VL on LRS-GRO and transfer evaluation on XLRS-Bench, compressing crops at high resolution into compact latent memories recovers most or all of the accuracy benefit of uncompressed crops at a fraction of their context cost. Controlled content substitutions confirm that model predictions rely on the encoded regional visual evidence rather than added token capacity, while restoring source crop pixels within the same model demonstrates that evidence availability governs reliance on memory. These findings establish compact visual latents as an effective mechanism for retaining observed local evidence in UHR remote sensing. Looking ahead, whereas current evaluations focus on individual local regions, complex Earth observation tasks often require comparing and aggregating evidence across several crops within a vast scene, where retaining uncompressed tokens quickly becomes infeasible. Extending our compact memory framework and diagnostic protocol to settings with multiple crops, together with constructing new UHR benchmarks for reasoning across multiple retained regions, represents a promising direction for future work.

{
    \small
    \bibliographystyle{ieeenat_fullname}
    \bibliography{main}

@inproceedings{alayrac2022flamingo,
  author = {Alayrac, Jean-Baptiste and Donahue, Jeff and Luc, Pauline and Miech, Antoine and Barr, Iain and Hasson, Yana and Lenc, Karel and Mensch, Arthur and Millican, Katherine and Reynolds, Malcolm and Ring, Roman and Rutherford, Eliza and Cabi, Serkan and Han, Tengda and Gong, Zhitao and Samangooei, Sina and Monteiro, Marianne and Menick, Jacob L. and Borgeaud, Sebastian and Brock, Andy and Nematzadeh, Aida and Sharifzadeh, Sahand and Bi\'{n}kowski, Miko\l{}aj and Barreira, Ricardo and Vinyals, Oriol and Zisserman, Andrew and Simonyan, Kar\'{e}n},
  title = {{Flamingo}: a Visual Language Model for Few-Shot Learning},
  booktitle = {Advances in Neural Information Processing Systems},
  volume = {35},
  pages = {23716--23736},
  year = {2022},
  publisher = {Curran Associates, Inc.},
  doi = {10.52202/068431-1723},
  url = {https://proceedings.neurips.cc/paper_files/paper/2022/hash/960a172bc7fbf0177ccccbb411a7d800-Abstract-Conference.html},
}

@inproceedings{jaegle2021perceiver,
  author = {Jaegle, Andrew and Gimeno, Felix and Brock, Andy and Vinyals, Oriol and Zisserman, Andrew and Carreira, Joao},
  title = {{Perceiver}: General Perception with Iterative Attention},
  booktitle = {Proceedings of the 38th International Conference on Machine Learning},
  series = {Proceedings of Machine Learning Research},
  volume = {139},
  pages = {4651--4664},
  year = {2021},
  publisher = {PMLR},
  url = {https://proceedings.mlr.press/v139/jaegle21a.html},
}

@inproceedings{hu2022lora,
  author = {Hu, Edward J. and Shen, Yelong and Wallis, Phillip and Allen-Zhu, Zeyuan and Li, Yuanzhi and Wang, Shean and Wang, Lu and Chen, Weizhu},
  title = {{LoRA}: Low-Rank Adaptation of Large Language Models},
  booktitle = {International Conference on Learning Representations},
  year = {2022},
  url = {https://openreview.net/forum?id=nZeVKeeFYf9},
}

@inproceedings{bolya2023tome,
  author = {Bolya, Daniel and Fu, Cheng-Yang and Dai, Xiaoliang and Zhang, Peizhao and Feichtenhofer, Christoph and Hoffman, Judy},
  title = {Token Merging: Your {ViT} but Faster},
  booktitle = {International Conference on Learning Representations},
  year = {2023},
  url = {https://openreview.net/forum?id=JroZRaRw7Eu},
}

@inproceedings{chen2024fastv,
  author = {Chen, Liang and Zhao, Haozhe and Liu, Tianyu and Bai, Shuai and Lin, Junyang and Zhou, Chang and Chang, Baobao},
  title = {An Image is Worth 1/2 Tokens After Layer 2: Plug-and-Play Inference Acceleration for Large Vision-Language Models},
  booktitle = {Computer Vision -- ECCV 2024},
  series = {Lecture Notes in Computer Science},
  volume = {15139},
  pages = {19--35},
  year = {2024},
  doi = {10.1007/978-3-031-73004-7_2},
  url = {https://www.ecva.net/papers/eccv_2024/papers_ECCV/html/10478_ECCV_2024_paper.php},
}

@inproceedings{he2024malmm,
  author = {He, Bo and Li, Hengduo and Jang, Young Kyun and Jia, Menglin and Cao, Xuefei and Shah, Ashish and Shrivastava, Abhinav and Lim, Ser-Nam},
  title = {{MA-LMM}: Memory-Augmented Large Multimodal Model for Long-Term Video Understanding},
  booktitle = {Proceedings of the IEEE/CVF Conference on Computer Vision and Pattern Recognition},
  pages = {13504--13514},
  year = {2024},
  url = {https://openaccess.thecvf.com/content/CVPR2024/html/He_MA-LMM_Memory-Augmented_Large_Multimodal_Model_for_Long-Term_Video_Understanding_CVPR_2024_paper.html},
}

@inproceedings{song2024moviechat,
  author = {Song, Enxin and Chai, Wenhao and Wang, Guanhong and Zhang, Yucheng and Zhou, Haoyang and Wu, Feiyang and Chi, Haozhe and Guo, Xun and Ye, Tian and Zhang, Yanting and Lu, Yan and Hwang, Jenq-Neng and Wang, Gaoang},
  title = {{MovieChat}: From Dense Token to Sparse Memory for Long Video Understanding},
  booktitle = {Proceedings of the IEEE/CVF Conference on Computer Vision and Pattern Recognition},
  pages = {18221--18232},
  year = {2024},
  url = {https://openaccess.thecvf.com/content/CVPR2024/html/Song_MovieChat_From_Dense_Token_to_Sparse_Memory_for_Long_Video_CVPR_2024_paper.html},
}

@article{bai2025qwen25vl,
  author = {Bai, Shuai and Chen, Keqin and Liu, Xuejing and Wang, Jialin and Ge, Wenbin and Song, Sibo and Dang, Kai and Wang, Peng and Wang, Shijie and Tang, Jun and Zhong, Humen and Zhu, Yuanzhi and Yang, Mingkun and Li, Zhaohai and Wan, Jianqiang and Wang, Pengfei and Ding, Wei and Fu, Zheren and Xu, Yiheng and Ye, Jiabo and Zhang, Xi and Xie, Tianbao and Cheng, Zesen and Zhang, Hang and Yang, Zhibo and Xu, Haiyang and Lin, Junyang},
  title = {{Qwen2.5-VL} Technical Report},
  journal = {arXiv preprint arXiv:2502.13923},
  year = {2025},
  doi = {10.48550/arXiv.2502.13923},
  url = {https://arxiv.org/abs/2502.13923},
}

@article{bai2025qwen3vl,
  author = {Bai, Shuai and Cai, Yuxuan and Chen, Ruizhe and Chen, Keqin and Chen, Xionghui and Cheng, Zesen and Deng, Lianghao and Ding, Wei and Gao, Chang and Ge, Chunjiang and Ge, Wenbin and Guo, Zhifang and Huang, Qidong and Huang, Jie and Huang, Fei and Hui, Binyuan and Jiang, Shutong and Li, Zhaohai and Li, Mingsheng and Li, Mei and Li, Kaixin and Lin, Zicheng and Lin, Junyang and Liu, Xuejing and Liu, Jiawei and Liu, Chenglong and Liu, Yang and Liu, Dayiheng and Liu, Shixuan and Lu, Dunjie and Luo, Ruilin and Lv, Chenxu and Men, Rui and Meng, Lingchen and Ren, Xuancheng and Ren, Xingzhang and Song, Sibo and Sun, Yuchong and Tang, Jun and Tu, Jianhong and Wan, Jianqiang and Wang, Peng and Wang, Pengfei and Wang, Qiuyue and Wang, Yuxuan and Xie, Tianbao and Xu, Yiheng and Xu, Haiyang and Xu, Jin and Yang, Zhibo and Yang, Mingkun and Yang, Jianxin and Yang, An and Yu, Bowen and Zhang, Fei and Zhang, Hang and Zhang, Xi and Zheng, Bo and Zhong, Humen and Zhou, Jingren and Zhou, Fan and Zhou, Jing and Zhu, Yuanzhi and Zhu, Ke},
  title = {{Qwen3-VL} Technical Report},
  journal = {arXiv preprint arXiv:2511.21631},
  year = {2025},
  doi = {10.48550/arXiv.2511.21631},
  url = {https://arxiv.org/abs/2511.21631},
}

@inproceedings{guo2026beyond,
  author    = {Guo, Jiawei and Chen, Yu and Wang, Xiang and Li, Shuai and Zhao, Xinpei and Liu, Huaxing and Dong, Shuai and Zhai, Feifei and Zhou, Yu},
  title     = {Beyond Visual Memory: Mechanistic Diagnostics of Latent Visual Reasoning},
  booktitle = {Proceedings of the 2026 Conference on Empirical Methods in Natural Language Processing},
  year      = {2026},
  pages     = {6493--6519},
  publisher = {Association for Computational Linguistics},
  doi       = {10.18653/2026.emnlp-main.284},
  url       = {https://aclanthology.org/2026.emnlp-main.284/}
}

@inproceedings{li2026imagination,
  author    = {Li, You and Chen, Chi and Li, Yanghao and Zeng, Fanhu and Huang, Kaiyu and Xu, Jinan and Sun, Maosong},
  title     = {Imagination Helps Visual Reasoning, But Not Yet in Latent Space},
  booktitle = {Proceedings of the 43rd International Conference on Machine Learning},
  year      = {2026},
  volume    = {306},
  series    = {Proceedings of Machine Learning Research},
  pages     = {66255--66269},
  publisher = {PMLR},
  url       = {https://proceedings.mlr.press/v306/li26r.html}
}

@inproceedings{viveiros2026holding,
  author    = {Viveiros, Andr{\'e} G. and Gon{\c{c}}alves, Nuno and Martins, Andr{\'e} F. T. and Lindemann, Matthias},
  title     = {What's Holding Back Latent Visual Reasoning?},
  booktitle = {Advances in Neural Information Processing Systems},
  year      = {2026},
  url       = {https://arxiv.org/abs/2605.18445},
  eprint    = {2605.18445},
  archivePrefix = {arXiv},
  primaryClass = {cs.CV}
}

@inproceedings{li2026latent,
  author = {Li, Bangzheng and Sun, Ximeng and Liu, Jiang and Wang, Ze and Wu, Jialian and Yu, Xiaodong and Chen, Hao and Barsoum, Emad and Chen, Muhao and Liu, Zicheng},
  title = {Latent Visual Reasoning},
  booktitle = {International Conference on Learning Representations},
  year = {2026},
  url = {https://proceedings.iclr.cc/paper_files/paper/2026/hash/efa0939caa8a906fd0a61a9c60e46e8b-Abstract-Conference.html}
}

@inproceedings{wang2026monet,
  author = {Wang, Qixun and Shi, Yang and Wang, Yifei and Zhang, Yuanxing and Wan, Pengfei and Gai, Kun and Ying, Xianghua and Wang, Yisen},
  title = {{Monet}: Reasoning in Latent Visual Space Beyond Image and Language},
  booktitle = {Proceedings of the IEEE/CVF Conference on Computer Vision and Pattern Recognition},
  year = {2026},
  pages = {12030--12040},
  url = {https://openaccess.thecvf.com/content/CVPR2026/html/Wang_Monet_Reasoning_in_Latent_Visual_Space_Beyond_Image_and_Language_CVPR_2026_paper.html}
}

@article{park2026cvrr,
  author = {Park, Suhyeong and Jung, Junha and Kang, Jaewoo},
  title = {Reason Through the Latent! Making Latent Visual Reasoning Necessary},
  journal = {arXiv preprint arXiv:2609.06746},
  year = {2026},
  doi = {10.48550/arXiv.2609.06746},
  url = {https://arxiv.org/abs/2609.06746}
}

@inproceedings{dong2026ilvr,
  author = {Dong, Shuai and Wang, Siyuan and Liu, Xingyu and Li, Chenglin and Hou, Haowen and Wei, Zhongyu},
  title = {Interleaved Latent Visual Reasoning with Selective Perceptual Modeling},
  booktitle = {Proceedings of the 64th Annual Meeting of the Association for Computational Linguistics (Volume 1: Long Papers)},
  year = {2026},
  pages = {29316--29335},
  publisher = {Association for Computational Linguistics},
  doi = {10.18653/v1/2026.acl-long.1351},
  url = {https://aclanthology.org/2026.acl-long.1351/}
}

@inproceedings{yang2026mirage,
  author = {Yang, Zeyuan and Yu, Xueyang and Chen, Delin and Shen, Maohao and Gan, Chuang},
  title = {Machine Mental Imagery: Empower Multimodal Reasoning with Latent Visual Tokens},
  booktitle = {Proceedings of the IEEE/CVF Conference on Computer Vision and Pattern Recognition},
  year = {2026},
  pages = {33510--33520},
  url = {https://openaccess.thecvf.com/content/CVPR2026/html/Yang_Machine_Mental_Imagery_Empower_Multimodal_Reasoning_with_Latent_Visual_Tokens_CVPR_2026_paper.html}
}

@article{wang2026geolens,
  author = {Wang, Fengxiang and Huang, Jiangnan and Chen, Mingshuo and Li, Yueying and Shi, Yang and Luo, Junwei and Wang, Haoyu and Li, Yansheng and Zhang, Jing and Zhao, Haiyan and Yang, Wenjing},
  title = {Beyond Zooming: Learning Multi-Tool Visual Reasoning for Ultra-High-Resolution Remote Sensing},
  journal = {arXiv preprint arXiv:2607.25993},
  year = {2026},
  eprint = {2607.25993},
  archivePrefix = {arXiv},
  url = {https://arxiv.org/abs/2607.25993}
}

@inproceedings{liu2026zoomearth,
  author = {Liu, Ruixun and Fu, Bowen and Song, Jiayi and Li, Kaiyu and Li, Wanchen and Xue, Lanxuan and Qiao, Hui and Zhang, Weizhan and Meng, Deyu and Cao, Xiangyong},
  title = {{ZoomEarth}: Active Perception for Ultra-High-Resolution Geospatial Vision-Language Tasks},
  booktitle = {Proceedings of the IEEE/CVF Conference on Computer Vision and Pattern Recognition (CVPR)},
  month = jun,
  year = {2026},
  pages = {34877--34888},
  url = {https://openaccess.thecvf.com/content/CVPR2026/html/Liu_ZoomEarth_Active_Perception_for_Ultra-High-Resolution_Geospatial_Vision-Language_Tasks_CVPR_2026_paper.html}
}

@inproceedings{wang2025xlrsbench,
  author = {Wang, Fengxiang and Wang, Hongzhen and Guo, Zonghao and Wang, Di and Wang, Yulin and Chen, Mingshuo and Ma, Qiang and Lan, Long and Yang, Wenjing and Zhang, Jing and Liu, Zhiyuan and Sun, Maosong},
  title = {{XLRS-Bench}: Could Your Multimodal {LLMs} Understand Extremely Large Ultra-High-Resolution Remote Sensing Imagery?},
  booktitle = {Proceedings of the IEEE/CVF Conference on Computer Vision and Pattern Recognition (CVPR)},
  month = jun,
  year = {2025},
  pages = {14325--14336},
  url = {https://openaccess.thecvf.com/content/CVPR2025/html/Wang_XLRS-Bench_Could_Your_Multimodal_LLMs_Understand_Extremely_Large_Ultra-High-Resolution_Remote_CVPR_2025_paper.html}
}

@inproceedings{luo2025lrsvqa,
  author = {Luo, Junwei and Zhang, Yingying and Yang, Xue and Wu, Kang and Zhu, Qi and Liang, Lei and Chen, Jingdong and Li, Yansheng},
  title = {When Large Vision-Language Model Meets Large Remote Sensing Imagery: Coarse-to-Fine Text-Guided Token Pruning},
  booktitle = {Proceedings of the IEEE/CVF International Conference on Computer Vision (ICCV)},
  month = oct,
  year = {2025},
  pages = {9206--9217},
  url = {https://openaccess.thecvf.com/content/ICCV2025/html/Luo_When_Large_Vision-Language_Model_Meets_Large_Remote_Sensing_Imagery_Coarse-to-Fine_ICCV_2025_paper.html}
}
}

\end{document}